\documentclass[conference]{IEEEtran}
\IEEEoverridecommandlockouts
\usepackage{cite}
\usepackage{amsmath,amssymb,amsfonts}
\usepackage{algorithmic}
\usepackage{graphicx}
\usepackage{textcomp}
\usepackage{xcolor}
\def\BibTeX{{\rm B\kern-.05em{\sc i\kern-.025em b}\kern-.08em
    T\kern-.1667em\lower.7ex\hbox{E}\kern-.125emX}}
    
\usepackage[setbb,setcolon]{kmath}

\usepackage[sort&compress,numbers]{natbib}

\usepackage{hyperref}

\usepackage{etoolbox}
\providetoggle{techreport}
\settoggle{techreport}{true}

\providetoggle{cameraready}
\settoggle{cameraready}{true}

\providetoggle{arxiv}
\settoggle{arxiv}{true}

\hypersetup{
  colorlinks=false,
  frenchlinks=false,
  pdfborder={0 0 0},
  naturalnames=false,
  hypertexnames=false,
}

\usepackage[T1]{fontenc}
\usepackage[utf8]{inputenc}
\usepackage[capitalize,noabbrev]{cleveref} % gestion intelligente des references
\crefname{equation}{equation}{equations}
\usepackage{amssymb}
\usepackage{amsmath}
\usepackage{amsfonts}
\usepackage{mathrsfs}
\usepackage{amsthm}
\usepackage{soul}		         % utilisé dans les macros de Charles

\usepackage{graphicx}
\usepackage{algorithmic}
\usepackage[algoruled, linesnumbered, onelanguage]{algorithm2e}
\def\u{{\mathbf u}}

\def\x{{\mathbf x}}
\def\y{{\mathbf y}}

\def\0{{\mathbf 0}}
\def\1{{\mathbf 1}}

\def\A{{\mathbf A}}

\def\tt{{\boldsymbol t}}

\def\Gc{{\mathcal G}}

\newtheorem{lemma}{Lemma}
\newtheorem{theorem}{Theorem}

\iftoggle{cameraready}{
	\newcommand{\suppCH}[1]{}
	\newcommand{\remCH}[1]{}

	\newcommand{\suppCE}[1]{}
	\newcommand{\remCE}[1]{}
	\newcommand{\addCE}[1]{#1}

	\newcommand{\suppCL}[1]{}
	\newcommand{\remCL}[1]{}
	\newcommand{\addCL}[1]{#1}
}{
	\def\suppCH#1{{\footnotesize \setstcolor{blue}\st{#1}}}
	\def\remCH#1{{\noindent\color{blue}{{\footnotesize [CH: #1]}}}}

	\newcommand{\suppCE}[1]{{\footnotesize \setstcolor{purple}\st{#1}}}
	\newcommand{\remCE}[1]{{\noindent\color{purple}{{\footnotesize [CE: #1]}}}}
	\newcommand{\addCE}[1]{{\noindent\color{purple}{#1}}}

	\newcommand{\suppCL}[1]{{\footnotesize \setstcolor{red}\st{#1}}}
	\newcommand{\remCL}[1]{{\noindent\color{red}{{\footnotesize [CL: #1]}}}}
	\newcommand{\addCL}[1]{{\noindent\color{red}{#1}}}
}

\newcommand{\paperfigs}[1]{\iftoggle{arxiv}{#1}{figs/#1}}
\newcommand{\codepath}{https://gitlab.inria.fr/cherzet/holder-safe}

\DeclareMathOperator{\Radius}{\normalfont{\texttt{Rad}}}

\def\ie{\textit{i.e.},\xspace}
\def\eg{\textit{e.g.},\xspace}
\def\etal{\textit{et al.}\xspace }

\newcommand{\indcatorfunc}{\eta}

\newcommand{\obsletter}{y}
\newcommand{\obs}{\mathbf{\obsletter}}

\newcommand{\obsdim}{m}

\newcommand{\dicomat}{\mathbf{A}}
\newcommand{\atom}{\mathbf{a}}

\newcommand{\pvletter}{x} 								% primal letter 
\newcommand{\pv}{\mathbf{\pvletter}}				% primal vector 
\newcommand{\pvdim}{n}									% dimension of primal vector
\newcommand{\pvopt}{\pv^\star} 						% primal solution

\newcommand{\idxpv}{i}

\newcommand{\primalfun}{P}								% primal objective function
\newcommand{\dualfun}{D}								% dual objective function
\newcommand{\LASSO}{Lasso}

\newcommand{\dvletter}{u}								% dual letter 
\newcommand{\dv}{\mathbf{\dvletter}}				% dual vector 
\newcommand{\dvopt}{\dv^\star}						% dual solution 
\newcommand{\dfset}{\mathcal{U}}						% dual feasible set

\newcommand{\dualscal}[1]{\textnormal{\texttt{DS}}}	% dual feasible set

\newcommand{\dpoint}{\dv_0}

\newcommand{\saferegion}{\mathcal{R}}
\newcommand{\someset}{\mathcal{S}}

\newcommand{\spheresymb}{\mathcal{B}}
\newcommand{\spherer}{R}    % Radius
\newcommand{\spherec}{\mathbf{c}}     % Center

\newcommand{\GapBallname}{{\textnormal{\texttt{gap}}}}
\newcommand{\GapBall}{ \spheresymb_{\GapBallname}}

\newcommand{\halfspacesymb}{\mathcal{H}}
\newcommand{\halfspacen}{\mathbf{g}}
\newcommand{\halfspacetreshold}{\delta}

\newcommand{\GapHalfspacename}{{\textnormal{\texttt{gap}}}}
\newcommand{\GapHalfspace}{\halfspacesymb_{\GapHalfspacename}}

\newcommand{\newHalfspace}{\halfspacesymb_{\textnormal{\texttt{new}}}}

\newcommand{\domesymb}{\mathcal{D}}
\newcommand{\GapDome}{\domesymb_{\textnormal{\texttt{gap}}}}
\newcommand{\NewDome}{\domesymb_{\textnormal{\texttt{new}}}}

\newcommand{\gapfun}{\mathrm{gap}}

\newcommand{\norm}[1]{\left\lVert#1\right\rVert}
\newcommand{\dotp}[2]{\langle #1, #2\rangle}
\begin{document}

% \title{New Lasso's Safe Screening Method Based on Supporting Hyperplane of Dual Feasible Set\\
% \title{Lasso Safe Screening Based on Supporting Hyperplane of Dual Feasible Set\\
% \title{Beyond GAP Screening: New dome test for Lasso
%\title{Beyond \LASSO GAP Screening: \\ introducing the \proposeddome{}
%\title{Beyond GAP Screening for \LASSO\\  with ``Hölder'' supporting hyperplanes
%\title{Beyond GAP Screening for \LASSO\\  with dual safe half-planes
\title{Beyond GAP screening for \LASSO\\  by exploiting new dual cutting half-spaces \\
	with supplementary material
	%based on estimated observations
	% {\footnotesize \textsuperscript{*}Note: Sub-titles are not captured in Xplore and
	% should not be used}
	% \thanks{Identify applicable funding agency here. If none, delete this.}
}

% \author{\IEEEauthorblockN{Thu Le Tran}
% 	\IEEEauthorblockA{
% 		\textit{Univ Rennes, INRIA-SMISMART , CNRS, IRMAR - UMR 6625, F-35000 Rennes, France} \\
% 		Rennes, France \\
% 		\href{mailto:thu-le.tran@univ-rennes1.fr}{you@univrennes1.fr}
% 	}
% 	\and
% 	\IEEEauthorblockN{Clément Elvira}
% 	\IEEEauthorblockA{
% 		\textit{IETR UMR CNRS 6164,
% 				CentraleSupelec Rennes Campus} \\
% 		35576 Cesson Sévigné, France \\
% 		\href{mailto:clement.elvira@centralesupelec.fr}{clement.elvira@centralesupelec.fr}
% 	}
% 	\and
% 	\IEEEauthorblockN{Cédric Herzet}
% 	\IEEEauthorblockA{\textit{INRIA Rennes - Bretagne Atlantique} \\
% 		Rennes, France \\
% 		cedric.herzet@inria.fr}
% 	\and
% 	\IEEEauthorblockN{Hong Phuong Dang}
% 	\IEEEauthorblockA{\textit{dept. name of organization (of Aff.)} \\
% 		City, Country \\
% 		email address or ORCID}	
% 	% \and
% 	% \IEEEauthorblockN{5\textsuperscript{th} Given Name Surname}
% 	% \IEEEauthorblockA{\textit{dept. name of organization (of Aff.)} \\
% 	% 	City, Country \\
% 	% 	email address or ORCID}
% }

\author{
	\IEEEauthorblockN{
		Thu-Le Tran\IEEEauthorrefmark{1},
		Cl\'ement Elvira\IEEEauthorrefmark{2},
		Hong-Phuong Dang\IEEEauthorrefmark{3}, and
		C\'edric Herzet\IEEEauthorrefmark{4}
	} \\
	% \vspace{0.5cm}

	\IEEEauthorblockA{\IEEEauthorrefmark{1}Univ Rennes, IRMAR - UMR 6625, F-35000 Rennes, France
		% \\ \href{mailto:thu-le.tran@univ-rennes1.fr}{thu-le.tran@univ-rennes1.fr}
	}%
	\IEEEauthorblockA{\IEEEauthorrefmark{2}IETR UMR CNRS 6164,			CentraleSupelec Rennes Campus, 35576 Cesson Sévigné, France
		% \\ \href{mailto:clement.elvira@centralesupelec.fr}{clement.elvira@centralesupelec.fr}
	}%
	\IEEEauthorblockA{\IEEEauthorrefmark{3}LaTIM, INSERM-UMR1101, {\itshape Univ. de Bretagne Occidentale}, Brest, France
		% \\ \href{mailto:dang@univ-brest.fr}{dang@univ-brest.fr}
	}% <-this % stops an unwanted space
	\IEEEauthorblockA{\IEEEauthorrefmark{4}Inria Rennes - Bretagne Atlantique, Rennes, France
		\\
		emails:
		\href{mailto:thu-le.tran@univ-rennes1.fr}{thu-le.tran@univ-rennes1.fr},
		\href{mailto:clement.elvira@centralesupelec.fr}{clement.elvira@centralesupelec.fr},
		\href{mailto:dang@univ-brest.fr}{dang@univ-brest.fr},
		\href{mailto:cedric.herzet@inria.fr}{cedric.herzet@inria.fr}
	}

	\thanks{
		\addCE{
			The research presented in this paper is reproducible.
			Code is available at \protect\url{\codepath}.
		}
	}

}

\maketitle

%!TEX root = ./paper.tex
%!TEX spellcheck = en_US

\begin{abstract}
	%%%%%%%%%%%%%%%%%%%%%%%%%%%%
	%   Introduce Lasso, Screening, New Dome and results
	%%%%%%%%%%%%%%%%%%%%%%%%%%%% 
	%\remCE{à réécrire à la fin}
	%Sparse representations have demonstrated a great potential in statistics, machine learning and signal processing, in which the task is to approximate an observation vector as a combination of a small number of known vectors called atoms.
	In this paper, we propose a novel safe screening test for \LASSO{}. 
	Our procedure is based on a safe region with a dome geometry
	and 
	%The construction of our dome 
	exploits a canonical representation of the set of half-spaces 
	(referred to as ``dual cutting half-spaces'' in this paper)
	%\remCL{Should we remove parentheses in abstract?}
	containing the dual feasible set. %of \LASSO. 
	The proposed safe region is shown to be always included in the state-of-the-art ``GAP Sphere'' and ``GAP Dome'' proposed by Fercoq \etal (and strictly so under very mild conditions)
	%\remCL{is it worth mentioning, or should we just ignore it?})
	while involving the same computational burden.   
	Numerical experiments confirm that our new dome enables to devise more powerful screening tests than  GAP regions and lead to significant acceleration to solve \LASSO.\\
	%in various setup as commpared to screening tests with our dome instead of GAP regions may reduce the complexity in various cases.
	% This task can be achieved by solving the so-called \LASSO problem.  using convex optimization algorithms and further improved by screening techniques which can safely reject all irrelevant atoms before/during the solving processes.
	%This note addresses a new safe screening method using dome region. Our NEW dome mainly relies on a cutting hyperplane normal to the iterative estimation of the observation.
	%In a comparison with GAP Sphere and GAP Dome, two state-of-the-art safe regions, we show that our dome is a proper subset of these regions with radius converges to zero $\sqrt{2}$-times faster. Finally, numerical experiments show that screening with our dome instead of GAP regions may reduce the complexity in various cases.
\end{abstract}


% \tableofcontents

%!TEX root = ./paper.tex
%!TEX spellcheck = en_US

\section{Introduction} \label{sec:introduction}

Finding sparse representations is a fundamental problem in signal processing and machine learning.
It consists in decomposing
some vector
\(\obs\in\kR^\obsdim\) as a linear combination of a few columns (referred to as \emph{atoms}) of a matrix \(\dicomat=[\atom_1,\dotsc,\atom_\pvdim]\in\kR^{\obsdim\times\pvdim}\) called \emph{dictionary}.
A popular strategy for obtaining sparse representations is to solve the so-called \LASSO{}  problem:
\begin{equation} \label{eq: primal}
	\pvopt\in \kargmin_{\pv \in {\kR^\pvdim}}
	\
	\primalfun(\pv) \triangleq \tfrac{1}{2}\kvvbar{\obs-\dicomat\pv}_2^2 +\lambda \kvvbar{\pv}_1
\end{equation}
where \(\lambda>0\), see \cite{Foucart2013book}.
Inasmuch as solving~\eqref{eq: primal} may require a heavy computational burden as \(\pvdim\) becomes large, the design of efficient optimization techniques tackling this problem has become an active field of research.
Among the most popular approaches addressing~\eqref{eq: primal}, one can mention~\cite{Figueiredo2007Gradient,beck2009fast,Boyd2011Distributed}.

A noteworthy approach in this field is the acceleration method proposed by El Ghaoui \etal in~\cite{Laurent-El-Ghaoui:2012qs} and known as \emph{safe screening},
%It consists in designing
%\remCL{Phrase "It consists in designing" is repeated}
which aims to design
simple tests to identify zeros entries in the minimizers of~\eqref{eq: primal}.
This knowledge can then be exploited to (potentially significantly) reduce the dimensionality of the problem by discarding the atoms of the dictionary weighted by zero.
Over the past few years, safe screening has sparked a surge of interest in the literature, see \textit{e.g.},~\cite{xiang2011learning,bonnefoy2014dynamic,fercoq2015mind,Herzet16Screening,Herzet:2019fj,dai2012ellipsoid,xiang2012fast,icml2014c2_liuc14} and beyond \(\ell_1\)-regularization~\cite{ndiaye2017gap,elvira2020safe,elvira2021safe,guyard2021screen}.

Standard screening methodologies leverage the concept of ``\textit{safe region}'', a set provably containing the optimal solution of the dual problem of \eqref{eq: primal}, see \eg \cite[Section 4]{Xiang:2017ty}.
The choice of the safe region reveals to be crucial to the final effectiveness and efficiency of the screening tests.
On the one hand, loosely speaking, ``smaller'' regions lead to more effective tests, see \cite[Lemma 1]{Xiang:2017ty}.
On the other hand, the complexity of the tests is closely related to the geometry of the safe region.
As a consequence, safe regions with ``simple'' geometries such as spheres \cite{Laurent-El-Ghaoui:2012qs,xiang2011learning,bonnefoy2014dynamic,fercoq2015mind,Herzet16Screening} or domes~\cite{xiang2012fast,icml2014c2_liuc14} are commonly considered in the literature.

One state-of-the-art methodology to find a good compromise between these two requirements was proposed in \cite{fercoq2015mind}:
%In this paper, 
the authors introduced two new safe regions (referred to as \textit{``GAP sphere''} and \textit{``GAP dome''}) whose radii are proportional to the duality gap attained by the primal-dual feasible couple used to design the region.
The radii of GAP regions have therefore the desirable feature to converge to zero when the primal-dual feasible couple tends to a primal-dual solution, and thus lead to extremely effective screening tests.

In this paper, we propose a new safe dome which is provably contained in the GAP regions.
% \remCL{remove "proposed in \cite{fercoq2015mind}" 
% as it is already mentioned above}.
The definition of our dome is based on a fine characterization of the set of half-spaces (referred to as ``\textit{dual cutting half-spaces}'' hereafter) containing the whole dual feasible set.
Its construction has the same complexity as GAP dome and also relies on
the identification of some primal-dual feasible couple.

The paper is organized as follows.
In the next section, we define the notations used throughout the paper.
In \Cref{sec:rappels Safe screening Lasso}, some background on \LASSO{} and safe screening is provided.
In \Cref{sec:holder dome}, we derive a fine characterization of the set of dual cutting half-spaces and present our new safe dome.
The relevance of our proposed approach is finally illustrated in \cref{sec:simu} via numerical simulations.

\section{Notations} \label{sec:notations}

% \remCE{À lisser}
We use the following notational conventions throughout the
paper.
Boldface uppercase (\textit{e.g.}, \(\dicomat\)) and lowercase (\textit{e.g.}, \(\pv\)) letters respectively represent matrices and vectors.
\(\0_{\pvdim}\) denotes the all-zeros vector of \(\kR^\pvdim\).
\(\dotp{\cdot}{\cdot}\) stands for the canonical inner product \addCE{between two vectors}.
% Since the dimension will usually be clear from the context, it is omitted in the notation.
%The \(\idxpv\)th column of a matrix \(\dicomat\) is denoted \(\atom\).
%Similarly, 
The \(\idxpv\)th entry of a vector \(\pv\) is denoted \(\pv(\idxpv)\).
Calligraphic letters (\textit{e.g.}, \(\mathcal{S}\)) are used for sets.
%\remCE{y a une ligne à gagner là. Retirer ``between two vectors?''}

%and the notation || refers to their cardinality.
% If S ⊆ {1, . . . , n} and x ∈ R n, xS denotes to the restriction of x to its elements indexed by S.
% Similarly, AS corresponds to the restriction of A ∈ R m×n to its columns indexed by S.

%!TEX root = ./paper.tex
%!TEX spellcheck = en_US

%\section{Safe screening for \LASSO} \label{sec:rappels Safe screening Lasso}
\section{Background} \label{sec:rappels Safe screening Lasso}

In this section, we provide
some elements of convex analysis for problem \eqref{eq: primal} and
recall
the main ingredients underlying the concept of safe screening for \LASSO.

\subsection{Dual problem and optimality conditions}
We first note that~\eqref{eq: primal} admits at least one minimizer
since \(\primalfun(\cdot)\) is continuous, proper and coercive~\cite[Theorem~2.14]{Beck2017aa}.
The dual problem of~\eqref{eq: primal} writes as
\begin{equation}\label{eq: dual}
	\dvopt = \kargmax_{\dv\in \dfset}
	\
	\dualfun(\dv)
	\triangleq
	\tfrac{1}{2}\kvvbar{\obs}_2^2 - \tfrac{1}{2} \kvvbar{\obs - \dv}_2^2
\end{equation}
where \(\dfset\triangleq \{\dv\in\kR^\obsdim : \Vert{\ktranspose{\dicomat}\dv}\Vert_\infty \leq \lambda\}\) is the so-called \emph{dual feasible set}, see \cite[Appendix A]{xiang2016screening}.
Since $\dfset$ is closed and \addCL{$\dualfun(\cdot)$} is strictly concave, problem~\eqref{eq: dual} admits a unique maximizer $\dvopt$.

It is known that strong duality holds between problems \eqref{eq: primal} and \eqref{eq: dual}, that is
\begin{align}
	\forall \pv\in\kR^\pvdim, \forall \dv\in\dfset:\
	\gapfun(\pv,\dv)\triangleq\primalfun(\pv)-\dualfun(\dv)\geq 0
\end{align}
with equality if and only if $(\pv,\dv)$ is primal-dual optimal, see \eg \cite[Theorem~A.2]{Beck2017aa}.
Moreover, any primal-dual optimal couple $(\pvopt,\dvopt)$ must verify the following optimality conditions \cite[Section~2]{xiang2016screening}:
\begin{equation}\label{eq: ux}
	% \stepcounter{equation}
	% \tag{\text{KKT}-\theequation}
	\u^\star=\y-\A\x^\star
\end{equation}
and,
%\remCH{J'enleve les tag speciaux pour le moment car on ne les utilise pas plus que d'aurtes equations}
\begin{equation}\label{eq: KKT}
	% \stepcounter{equation}
	% \tag{\text{KKT}-\theequation}
	\dotp{\atom_\idxpv}{\dvopt} =
	\begin{cases}
		\lambda\ \mathrm{sign}(\pvopt(\idxpv)) & \text{ if } \pvopt(\idxpv) \neq 0, \\
		s \in \kintervcc{-\lambda}{\lambda}    & \text{ otherwise.}
	\end{cases}
\end{equation}
for all \(\idxpv=1,\dotsc,\pvdim\).
It is easy to see from these conditions that $\pvopt={\bf0}_\pvdim$ is the unique solution of \eqref{eq: primal} if and only if
\begin{align}
	\label{eq:def lambda max}
	\lambda \geq \lambda_{\max}\triangleq \|\ktranspose{\dicomat}\obs\|_\infty. \\[-0.2cm]
	\nonumber
\end{align}
%\textcolor{red}{Rajouter \(\lambda < \lambda_{\max}\)?}

\subsection{Safe screening}
Safe screening tests leverage the following consequence of~\eqref{eq: KKT}:
\begin{equation} \label{eq: ideal safe screening rule}
	\kvbar{\dotp{\atom_\idxpv}{\dvopt}} < \lambda \;\Longrightarrow\; \pvopt(\idxpv)=0.
\end{equation}
%\remCL{We didn't explain why screening tests accelerate solving methods?}
%\remCE{Agree, although we more or less did it in the introduction. Unfortunately, It will be (currently) difficult to add something rigorous with the limit of page. Here is an already too long but unsatisfying proposition. Otherwise, maybe removing the note after the definition of the Half-space? If prefer keeping the note on the half space}
% \addCL{If \(\atom_\idxpv\) passes this test, the dimension of Lasso problem is reduced since the \(\idxpv\)th-column  of \(\A\) can be removed without changing \(\pvopt\).}
%<<<<<<< HEAD
If the inequality in the left-hand side of \eqref{eq: ideal safe screening rule} is verified for some index \(\idxpv\),
the corresponding column of \(\dicomat\) can therefore be safely removed without changing the minimum value of~\eqref{eq: primal}.  % \remCL{agree}.
%=======
%\addCE{If an entry \(\idxpv\) passes this test, the corresponding column of \(\dicomat\) can therefore be safely removed without changed the minimum value of~\eqref{eq: primal}.}
%>>>>>>> 1069d061cd75b9dc2ffa50775421498d648e9c37
Although computing \(\dvopt\) is (generally) as difficult as solving~\eqref{eq: primal}, a weaker version of \eqref{eq: ideal safe screening rule} can be obtained if a region \(\saferegion\subseteq\kR^\pvdim\) containing $\dvopt$ (often called ``safe region'') is known.
%Implication~
\eqref{eq: ideal safe screening rule} can then be relaxed to:
\begin{equation}
	\label{eq:relax screening rule}
	\max_{\dv \in \saferegion} \vert{\dotp{\atom_\idxpv}{\dv}}\vert < \lambda
	\;\Longrightarrow\; \pvopt(\idxpv) = 0
	.
\end{equation}
%From the point of view of effectiveness of the screening test,
From an effectiveness point of view,
%safe region 
$\saferegion$ should be chosen as small as possible. In particular, if $\saferegion\subseteq	\saferegion'$ then obviously
\begin{equation}\label{eq:implication tests}
	\max_{\dv \in \saferegion'} \vert{\dotp{\atom_\idxpv}{\dv}}\vert < \lambda
	\implies
	\max_{\dv \in \saferegion} \vert{\dotp{\atom_\idxpv}{\dv}}\vert < \lambda
	,
\end{equation}
that is the screening test built from $\saferegion$ will always detect at least as many zeros as that constructed with $\saferegion'$.

%From the point of view of complexity,
%<<<<<<< HEAD
From a complexity point of view,
the computational cost of \eqref{eq:relax screening rule} mostly depends on the evaluation of $\max_{\dv \in \saferegion} \vert{\dotp{\atom_\idxpv}{\dv}}\vert$.
%\addCL{the complexity of evaluating $\max_{\dv \in \saferegion} \vert{\dotp{\atom_\idxpv}{\dv}}\vert$}.
% =======
% \addCL{From the computing point of view},
% the computational cost of \eqref{eq:relax screening rule} mostly depends on
% \addCL{the complexity of evaluating $\max_{\dv \in \saferegion} \vert{\dotp{\atom_\idxpv}{\dv}}\vert$}.
% >>>>>>> c8b76b9890944712dd6d5771a0e2bc2320b1a4cd
%our ability to evaluate $\max_{\dv \in \saferegion} \vert{\dotp{\atom_\idxpv}{\dv}}\vert$ efficiently.
A simple strategy to lower the cost of this operation consists in building safe regions with ``appropriate'' geometries.
%\remCE{Je fais mon relou: en francais, j'aurai écris "géométrie bien choisie" -> ``appropriate'' en anglais?}
%From the point of view of efficiency, the complexity of screening test \eqref{eq:relax screening rule} depends on our ability to evaluate $\max_{\dv \in \saferegion} \vert{\dotp{\atom_\idxpv}{\dv}}\vert$ efficiently.
Two standard choices are spheres and domes.
%Two types of geometries have been mainly considered in the literature: spheres and domes.

Sphere regions are defined by their center $\spherec\in\kR^\obsdim$ and radius $\spherer\geq 0$:
\begin{align}
	\saferegion=\spheresymb(\spherec,\spherer) \triangleq\{ \dv\in \kR^\obsdim: \kvvbar{\dv-\spherec}_2 \leq \spherer\}
	.
\end{align}
Particularizing the left-hand side of~\eqref{eq:relax screening rule} to the case where $\saferegion$ is a sphere, we obtain:
\begin{equation}
	\max_{\dv\in \spheresymb(\spherec, \spherer)}\
	\kvbar{\dotp{\atom_\idxpv}{\dv}} = \kvbar{\dotp{\atom_\idxpv}{\spherec}} +\spherer\kvvbar{\atom_\idxpv}_2
	.
\end{equation}
We see that the solution of the maximization problem then admits a simple closed form. Its computation only requires the evaluation  of one inner product between $\atom_\idxpv$ and $\spherec$.

Another popular choice of geometry is dome region.
A dome is defined as the intersection of a sphere and an half-space, that is
\begin{align}\label{eq: general def dome as intersection sphere and halfspace}
	\saferegion =  \domesymb(\spherec,\spherer,\halfspacen,\halfspacetreshold) \triangleq\spheresymb(\spherec,\spherer) \cap\halfspacesymb(\halfspacen,\halfspacetreshold)
\end{align}
where \(\halfspacen\in\kR^\obsdim\), \(\halfspacetreshold\in\kR\) and\footnote{Note that when \(\halfspacen=\0_\obsdim\), our definition implies that \(\halfspacesymb(\halfspacen,\halfspacetreshold)\) either reduces to \(\kR^\obsdim\) if \(\halfspacetreshold\geq0\) or is empty if \(\halfspacetreshold<0\).}
\begin{align}
	\halfspacesymb(\halfspacen,\halfspacetreshold)
	\triangleq
	\kset{\dv\in\kR^\obsdim}{\dotp{\halfspacen}{\dv}\leq \halfspacetreshold}.
\end{align}
% \addCL{Here, we emphasize the fact that \(\halfspacesymb(\halfspacen,\halfspacetreshold)\) should be understood in the general sense, \ie \(\halfspacesymb(\halfspacen,\halfspacetreshold)\) is a unsual half-space if \(\halfspacen \neq \0_m\), otherwise, it can be the whole space \(\ObsSpace\) (\(\halfspacen=\0_m\) and \( \halfspacetreshold\geq 0\)) or the empty set (\(\halfspacen=\0_m\) and \( \halfspacetreshold<0\)).}
%\remCH{You are right Le, but I think there is no need to stress this fact due to space limitation.}
% 
In this case, the left-hand side of~\eqref{eq:relax screening rule} also admits a closed-form solution, see \cite[Lemma.~3]{xiang2016screening}.
%\suppCL{In particular}
More precisely,
we have
\begin{align}
	\max_{\dv\in\domesymb}|\dotp{\atom_\idxpv}{\dv}|=\max\kparen{\max_{\dv\in\domesymb}\ \dotp{\atom_\idxpv}{\dv},\max_{\dv\in\domesymb}\ \dotp{-\atom_\idxpv}{\dv}}
\end{align}
and \(\forall \atom \in \kR^\obsdim \setminus \{\0_m\}\), \(\domesymb\neq \emptyset\) \addCE{and}
%\remCL{this ensures \(\psi_2\geq -1\)}:
%\addCL{is the max of $\max_{\dv\in \domesymb} {\dotp{\pm \atom_\idxpv}{\dv}}$. Here $\max_{\dv\in \domesymb} {\dotp{\atom_\idxpv}{\dv}}$ (similarly for $\max_{\dv\in \domesymb} {\dotp{-\atom_\idxpv}{\dv}}$) can be evaluated in a closed form} \cite[Lem.~3]{xiang2016screening}:
\begin{equation}\label{eq: dome maximization}
	% \max_{\dv\in \domesymb} {\dotp{\atom_\idxpv}{\dv}} = \dotp{\atom_\idxpv}{\spherec}
	% +  \spherer \kvvbar{\atom_\idxpv}_2 f(\psi_1,\psi_2),
	\max_{\dv\in \domesymb} {\dotp{\atom}{\dv}} = \dotp{\atom}{\spherec}
	+  \spherer \kvvbar{\atom}_2 f(\psi_1,\psi_2),
\end{equation}
where
\begin{align}
	\nonumber
	f(\psi_1, \psi_2)
	 & =
	\begin{cases}
		1                                                 & \text{if } \psi_1 \leq \psi_2, \\
		\psi_1\psi_2 + \sqrt{1-\psi^2_1}\sqrt{1-\psi_2^2} & \text{otherwise}
	\end{cases} \\
	% \psi_1
	%  & = \frac{
	% 	\dotp{\atom_\idxpv}{\halfspacen}
	% }{
	% 	\kvvbar{\atom_\idxpv}_2\kvvbar{\halfspacen}_2
	% },
	\psi_1
	 & = \frac{
		\dotp{\atom}{\halfspacen}
	}{
		\kvvbar{\atom}_2\kvvbar{\halfspacen}_2
	},
	\quad
	\psi_2 =
	\min \left( \frac{
		\halfspacetreshold-\dotp{\halfspacen}{\spherec}
	}{
		\spherer\kvvbar{\halfspacen}_2
	}, 1\right)
	.
	\nonumber
\end{align}
% \remCH{A reflechir mais je pense qu'on peut alléger (et potentiellement enlever les def des $\psi_i$ en supposant $\|\halfspacen\|_2=1$ et $\|\atom_\idxpv\|_2=1$)}
% \remCE{Aggree, j'avais hésité aussi. Pour $\|\atom_\idxpv\|_2=1$ c'est clair. Pour $\|\halfspacen\|_2=1$, ca alourdit un peu les définitions des régions safes (car du coup à la fois \(\halfspacen\) et \(\halfspacetreshold\) sont impactés).}
% This result can be found in, \eg~\cite[Lem.~3]{xiang2016screening}.
% \addCL{Here, we assume implicitly that \(\spherer>0\) and \(\halfspacen\neq \0_m\), otherwise we can use the sphere test.}
% \remCE{I do not agree with ``we assume''; I suggest something in that spirit if you want to mention: ``Note that the dome test reduces to the sphere if \(sphere \subseteq dome\).''}
% \remCL{Agree}
%\remCH{Idem}
% \remCH{A reflechir mais je pense qu'on peut alléger (et potentiellement enlever les def des $\psi_i$ en supposant $\|\halfspacen\|_2=1$ et $\|\atom_\idxpv\|_2=1$)}
% \remCE{Aggree, j'avais hésité aussi. Pour $\|\atom_\idxpv\|_2=1$ c'est clair. Pour $\|\halfspacen\|_2=1$, ca alourdit un peu les définitions des régions safes (car du coup à la fois \(\halfspacen\) et \(\halfspacetreshold\) sont impactés).}
% This result can be found in, \eg~\cite[Lem.~3]{xiang2016screening}.
The cost associated to \eqref{eq: dome maximization} is mostly dominated by the computation of %the inner products 
$\dotp{\halfspacen}{\spherec}$, $\dotp{\atom_\idxpv}{\spherec}$ and $\dotp{\atom_\idxpv}{\halfspacen}$.
% \remCE{I would remove the \(\pm\) sign since, in practice, the inner product is evaluated once (you you flip each sign to otain ther other)}
% \remCL{Agree. Should we mention that?}
% \remCE{Nop, I suggest we keep to old version, without the ``\(\pm\)''}
% \remCL{Agree}
Dome tests are therefore usually slightly more complex to implement than sphere tests.
%However, since dome geometry also enables to obtain smaller safe regions and they also lead potentially to much more effective tests.
%\remCE{Sonne étrange à l'oreille. Proposition:}
However, since dome geometry enables smaller safe regions, they potentially lead to much more effective tests.

\subsection{GAP regions}\label{sec:GAP regions}

To conclude this section, we provide the expressions of two well-known state-of-the-art safe regions
proposed in \cite{fercoq2015mind}.
The first region takes the form of a sphere and is known as the \textit{``GAP sphere''} in the literature.
It is defined by the following choice of parameters:
\begin{align}
	% \GapBallc &= \dv\\
	% \GapBallr &= \sqrt{2(\primalfun(\pv)-\dualfun(\dv))})
	\spherec & = \dv                                                    \\
	%\spherer &= \sqrt{2(\primalfun(\pv)-\dualfun(\dv))})	\label{eq:GAP sphere radius}
	\spherer & = \sqrt{2 \gapfun(\pv,\dv)})	\label{eq:GAP sphere radius}
\end{align}
where \((\pv,\dv)\) can be any primal-dual feasible couple.
%sphere and dome regions. These regions have been proposed in \cite{fercoq2015mind} and read: 
The second has the geometry of a dome and is usually dubbed \textit{``GAP dome''}.
It is defined by the following set of parameters:
\begin{align}
	\spherec           & = \tfrac{1}{2}(\obs+\dv) \label{eq: gap dome: spherec}      \\
	\spherer           & = \tfrac{1}{2}\|\obs-\dv\|_2 \label{eq: gap dome: spherer}  \\
	\halfspacen        & = \obs - \spherec \label{eq: gap dome: halfspacen}          \\
	% \halfspacetreshold &= \dotp{\halfspacen}{\spherec}+ \primalfun(\pv)-\dualfun(\dv)-\spherer^2
	\halfspacetreshold & = \dotp{\halfspacen}{\spherec}+ \gapfun(\pv,\dv)-\spherer^2
\end{align}
% \begin{align}
% 	\begin{array}{lcl}
% 		\spherec = (\obs+\dv)/2		&	& \halfspacen = \obs - \spherec \\[0.2cm]
% 		\spherer = \|\obs-\dv\|_2/2 &   & \halfspacetreshold = \dotp{\halfspacen}{\spherec}+ \gapfun(\pv,\dv)-\spherer^2
% 	\end{array}
% \end{align}
where $(\pv,\dv)$ can be any primal-dual feasible couple.
In the sequel, with a slight abuse of notation, we will respectively denote the GAP sphere and the GAP dome as $\GapBall(\pv,\dv)$ and $\GapDome(\pv,\dv)$ to explicitly emphasize their dependence on the choice of the primal-dual feasible couple $(\pv,\dv)$.

Although not proved explicitly in \cite{fercoq2015mind}, it can be easily shown (see \iftoggle{techreport}{Appendix~\ref{proof:eq:inclusion sage GAP regions}}{\cite[Appendix~\ref{techreport-proof:eq:inclusion sage GAP regions}]{Le2022_HolderDomeTechreport}}) that
\begin{align}\label{eq:inclusion sage GAP regions}
	\GapDome(\pv,\dv)\subseteq\GapBall(\pv,\dv).
\end{align}
%Moreover, 
These two regions have the desirable feature of having a
%\suppCL{volume} \addCL{radius} 
radius
\addCL{(see \eqref{eq:def radius closed bounded set})}
%\remCL{radius =0 is "stronger" than volume=0} \remCE{Agree, but the radius is not defined so far}
%\remCH{Keep radius since we can assume that it is a standard notion?} 
that decreases to zero when the couple $(\pv,\dv)$ tends to a primal-dual solution. In particular,
% \addCL{
% \begin{align}\label{eq:opt GAP sphere}
% 	\{ \dvopt\}=\GapBall(\pvopt,\dvopt)
% \end{align}
% }
if $(\pv,\dv)=(\pvopt,\dvopt)$, $\{ \dvopt\}=\GapBall(\pvopt,\dvopt)$ and the screening tests based on GAP regions reduce to
%\addCL{optimal screening test}
\eqref{eq: ideal safe screening rule}.\\

\section{Safe dome with general dual cutting half-spaces} \label{sec:holder dome}

In this section, we present and motivate our new safe dome region.
%The construction of our dome is based on the canonical characterization of the set of dual cutting halspaces, that is the half-spaces $\halfspacesymb(\halfspacen,\halfspacetreshold)$ containing the dual feasible set $\dfset$.
The construction of our dome is based on the set of dual cutting half-spaces $\halfspacesymb(\halfspacen,\halfspacetreshold)$: %containing the dual feasible set $\dfset$: 
%Since $\halfspacen$, $\halfspacetreshold$ can always be mutiplied by a \addCH{positive} constant with no change in the definition of $\halfspacesymb(\halfspacen,\halfspacetreshold)$, we assume hereafter that $\halfspacetreshold=\lambda$.\remCH{I do not understand why you want to remove the previous sentence ? } 
%The next lemma provides a canonical characterization of the set% of \addCL{such $\halfspacen$}:
\begin{align}
	% \mathcal{G} = \kset{\halfspacen\in\kR^\obsdim}{|\dotp{\halfspacen}{\dv}|\leq \lambda\ \forall \dv\in\dfset}.
	\mathcal{G} \triangleq \kset{(\halfspacen,\halfspacetreshold)\in\kR^\obsdim\times\kR}{\dotp{\halfspacen}{\dv}\leq \halfspacetreshold\ \forall \dv\in\dfset}.
\end{align}
The next lemma provides a canonical characterization of \(\Gc\).
A proof can be found in Appendix~\ref{proof:lemma: canonical characterization}.
\begin{lemma}\label{lemma: canonical characterization}
	\begin{align}
		\mathcal{G} = \kset{(\dicomat\pv,\halfspacetreshold)}{\pv\in\kR^\pvdim,\halfspacetreshold\geq\lambda \|\pv\|_1}
		.
	\end{align}
\end{lemma}

%\remCL{Actually, we have a stronger result: $\dfset/\lambda$ and $\Gc$ are polar set of each other, see  \href{https://en.wikipedia.org/wiki/Bipolar_theorem}{Bipolar Theorem}.}

% In the sequel, we will refer to the
% half-spaces defined by
% elements of
% $\mathcal{G}$ as
% ``dual-set safe''
% (ou ``$\dfset$-safe'')
% half-spaces since they are ensured to contain the whole dual feasible set $\dfset$.
% \remCH{Faire reference aux cas particuliers ? 1) Single atom, 2) supporting hyperplane based on optimality conditions. }

%We now expose our proposed new dome region and further demonstrate that it is guaranteed to perform at least comparably in terms of performance detection compared to the two GAP-based regions.
We now expose our proposed new dome region and state a result showing that it is guaranteed to perform at least as well as GAP safe regions presented in Section~\ref{sec:GAP regions}.

The definition of the proposed dome is encapsulated in the following theorem:
\begin{theorem}\label{th:def and safeness new dome}
	Let \(\pv\in\kR^\pvdim\), \(\dv\in \dfset\) and
	\begin{align}
		\spherec           & = \tfrac{1}{2}(\obs+\dv)     \\
		\spherer           & = \tfrac{1}{2}\|\obs-\dv\|_2 \\
		\halfspacen        & = \dicomat\pv                \\
		\halfspacetreshold & = \lambda \|\pv\|_1
	\end{align}
	Then
	\begin{align}
		\dvopt\in\domesymb(\spherec,\spherer,\halfspacen,\halfspacetreshold).
	\end{align}
\end{theorem}
A proof of this result can be found in Appendix~\ref{proof:th:def and safeness new dome}.
% \textit{Proof:}
% First remember that a dome is defined as the intersection of the ball $\spheresymb(\spherec,\spherer)$ and the half-space $\halfspacesymb(\halfspacen,\halfspacesymb)$, see \eqref{eq: general def dome as intersection sphere and halfspace}. It is then sufficient to show that both $\spheresymb(\spherec,\spherer)$  and $\halfspacesymb(\halfspacen,\halfspacetreshold)$ are safe. 
% The safeness of $\spheresymb(\spherec,\spherer)$ is proved in \cite[Section 2.2]{icml2014c2_liuc14}. The safeness of $\halfspacesymb(\halfspacen,\halfspacesymb)$ follows 
% from Lemma~\ref{lemma: canonical characterization}. 
% 
The safeness of \(\halfspacesymb(\dicomat\pv,\lambda\|\pv\|_1)\) can in fact be seen as a simple consequence of the Hölder inequality:
\begin{align}\nonumber
	\langle\dicomat\pv, \dvopt\rangle=\langle\pv, \ktranspose{\dicomat}\dvopt\rangle\leq \|\pv\|_1\|\ktranspose{\dicomat}\dvopt\|_\infty \leq \lambda\|\pv\|_1,
\end{align}
where the last inequality follows from dual feasibility of \(\dvopt\).
%Finally, the safeness of $\spheresymb(\spherec,\spherer)$ and $\halfspacesymb(\halfspacen,\halfspacesymb)$ respectively follows from \cite[Section 2.2]{icml2014c2_liuc14} and Lemma~\ref{lemma: canonical characterization}.
% \hfill$\square$ \\
%
%
% \textit{Proof}:
% the proof of the result is straightfoward.
% First remember that a dome is defined as the intersection of the ball $\spheresymb(\spherec,\spherer)$ and the half-space $\halfspacesymb(\halfspacen,\halfspacesymb)$, see \eqref{eq: general def dome as intersection sphere and halfspace}. It is then sufficient to show that both $\spheresymb(\spherec,\spherer)$  and $\halfspacesymb(\halfspacen,\halfspacetreshold)$ are safe. Finally, the safeness of $\spheresymb(\spherec,\spherer)$ and $\halfspacesymb(\halfspacen,\halfspacesymb)$ respectively follows from \cite[Section 2.2]{icml2014c2_liuc14} Lemma~\eqref{lemma: canonical characterization}. \hfill$\square$\\
%
% Similarly to the GAP regions, the definition of the dome in Theorem~\ref{th:def and safeness new dome} only depends on the choice of a primal-dual feasile couple $(\pv,\dv)$. 
%
In~the sequel we will therefore refer to the dome defined in Theorem~\ref{th:def and safeness new dome} as ``\textit{Hölder dome}''.
% since the safeness of $\halfspacesymb(\dicomat\pv,\lambda\|\pv\|_1)$ can be seen as a direct consequence of the Hölder inequality.
We note that, similarly to the GAP regions, the Hölder dome is completely specified by the choice of a primal-dual feasible couple $(\pv,\dv)$.
Hereafter, we will use the notation $\NewDome(\pv, \dv)$ to emphasize this fact.

We note that the definition of $\NewDome(\pv, \dv)$ only differs from that of $\GapDome(\pv,\dv)$ in the choice of the half-space $\halfspacesymb(\halfspacen,\halfspacetreshold)$, where the canonical characterization of $\mathcal{G}$ in Lemma~\ref{lemma: canonical characterization} is directly exploited in the former. Our next result shows that this choice is beneficial regarding the size of the safe region:
\begin{theorem} \label{theo: subsets}
	\addCL{For} \(\pv\in\kR^\pvdim\),
	%\(\forall \pv\in\kR^\pvdim\),
	%\remCL{or $\A\x\neq \0_m$?}\remCE{At the end, where do we use this hypothesis?}
	\(\dv\in \dfset\):
	\begin{equation}\label{eq: theo eq1}
		%\dvopt \in 
		\NewDome(\pv,\dv)
		\subseteq \GapDome(\pv,\dv)
		%\subseteq \GapBall(\pv,\dv).
		.
	\end{equation}
	%\addCH{Moreover, the inclusion is strict if and only if $\primalfun(\pv)<\primalfun({\bf0}_\pvdim)$.}
	% \addCL{Moreover, the inclusion is strict if $\primalfun(\pv)<\primalfun({\bf0}_\pvdim)$.}
	Moreover, if \(\primalfun(\pv)<\primalfun({\bf0}_\pvdim)\) and \((\pv,\dv)\) is not primal-dual optimal, then the inclusion is strict.
	%\remCH{TO DO: Complete the proof, Le can you deal with it?}
	% \remCE{Is there an hypothesis that is missing? because we may have \(\NewDome(\pvopt,\dvopt)=\GapDome(\pvopt,\dvopt)\) right?}
	% \suppCH{In particular} \addCH{Moreover},
	% \addCL{\(\GapBall(\pv,\dv)=\{\dvopt\}\)}
	% \emph{if and only if} \(\dv=\dvopt\) and \(\pv\) is a minimizer of~\eqref{eq: primal}.
	% \remCE{Proof of second point not done}
\end{theorem}
%\remCL{I think that we don't need to assume that $\A\x=\0_m$	if we agree that the ``half-space'' can be understood in a general sense. Indeed, if $\A\x=\0_m$ (and/or $\x=\0_n$), then $\Hc(\A\x, \lambda||\x||_1)$ equals to the \emph{whole space} $\Rbb^m$. Therefore, Theorem 2 is still valid in such cases. Anyway, I agree that $\x=\0_n$ may break the terminology ``half-space'' strictly speaking.}
\noindent
%\remCL{I think that $\x\neq \0_n$ is a condition to apply Lemma 1, i.e. under the condition $\x\neq \0_n$ then Lemma 1 implies $\dotp{\frac{\A\x}{||x||_1}}{\u}\leq \lambda$. I hope that I understand your idea correctly?!}
% \remCH{I fixed Lemma 1. I think we can remove the assumption $\pv = {\bf0}_\pvdim$ now. }
% \remCL{Wao, I see and agree!}
A proof of this result can be found in Appendix~\ref{sec:app:proof theorem subsets}.
Theorem~\ref{theo: subsets} shows that the Hölder dome is guaranteed to be a subset of the GAP dome and sphere.
From~\eqref{eq:implication tests}, this suggests that screening tests based on our proposed dome are ensured to perform at least as well as the two other safe regions.
% \addCL{Note that, if \((\pv,\dv)\) is primal-dual optimal, then
% \(\{\dvopt\}=\NewDome(\pv,\dv)
% \subseteq \GapDome(\pv,\dv)
% \subseteq \GapBall(\pv,\dv)\).}
We also note that the condition ``$\primalfun(\pv)<\primalfun({\bf0}_\pvdim)$'', ensuring
% \remCE{Ensuring ``along with'' désormais. Je propose de laisser comme ca sinon il faudrait discuter ce que \((\pv,\dv)\neq(\pvopt,\dvopt)\) implique} the strict inclusion \suppCE{of} \remCE{?} 
\(\NewDome(\pv,\dv)\subset\GapDome(\pv,\dv)\),
%\remCE{Il manque les paramètres dans la définition?}
is very mild and verified in many practical setups.
It is for example the case when the iterates $\{\pv^{(t)}\}_{t=1}^\infty$ of an optimization procedure, initialized at $\pv^{(0)}={\bf0}_\pvdim$ and monotically decreasing the cost function, are used to define the primal point $\pv$ used in the construction of the dome.

\section{Numerical Experiment}\label{sec:simu}

This section reports an empirical study of the relevance of the Hölder dome presented in \Cref{sec:holder dome}.\footnote{
	%The research presented in this paper is reproducible.
	The code is available at \protect\url{\codepath}.
}
% compared to GAP-based screening tests.
% assess the gain in performances of the Hölder dome presented in \Cref{sec:holder dome} compared to GAP-based screening tests.
For the two experiments, our simulation setup is as follows.
We set \((\obsdim, \pvdim)=(100, 500)\).
% \remCE{\((100,500)\)?}
For each trial, new realizations of \(\dicomat\) and \(\obs\) are generated.
% for each trial.
The observation \(\obs\) is drawn according to a uniform distribution on the \(\obsdim\)-dimensional unit sphere while \(\dicomat\) either satisfies \textit{i)} the entries are i.i.d. realizations of a normal distribution or \textit{ii)} \(\dicomat\) has a Toeplitz structure, \ie  columns are shifted versions of a Gaussian curve.
The columns of \(\dicomat\) are then normalized such that \(\Vert\atom_\idxpv\Vert_2=1\) for all \(\idxpv\).
%\remCE{Discutable?}\remCH{Non, c'est standard}\remCE{Je voulais dire, comme on ne fait pas l'hypothèse dans le papier, on ne risque pas de nous le roprocher?}\remCH{Je ne pense pas. On n'est jamais à l'abris d'un débile toutefois ;)}
% Finally, we set \((m, n)=(100, 300)\) and consider several choices of ratio \(\lambda/\lambda_{\max}\) (see~\eqref{eq:def lambda max}).
%Note that we only con-
% Each results are averaged over \(100\) Monte-Carlo simulations.

\begin{figure}[t]
	\centering
	\includegraphics[width=.99\columnwidth]{\paperfigs{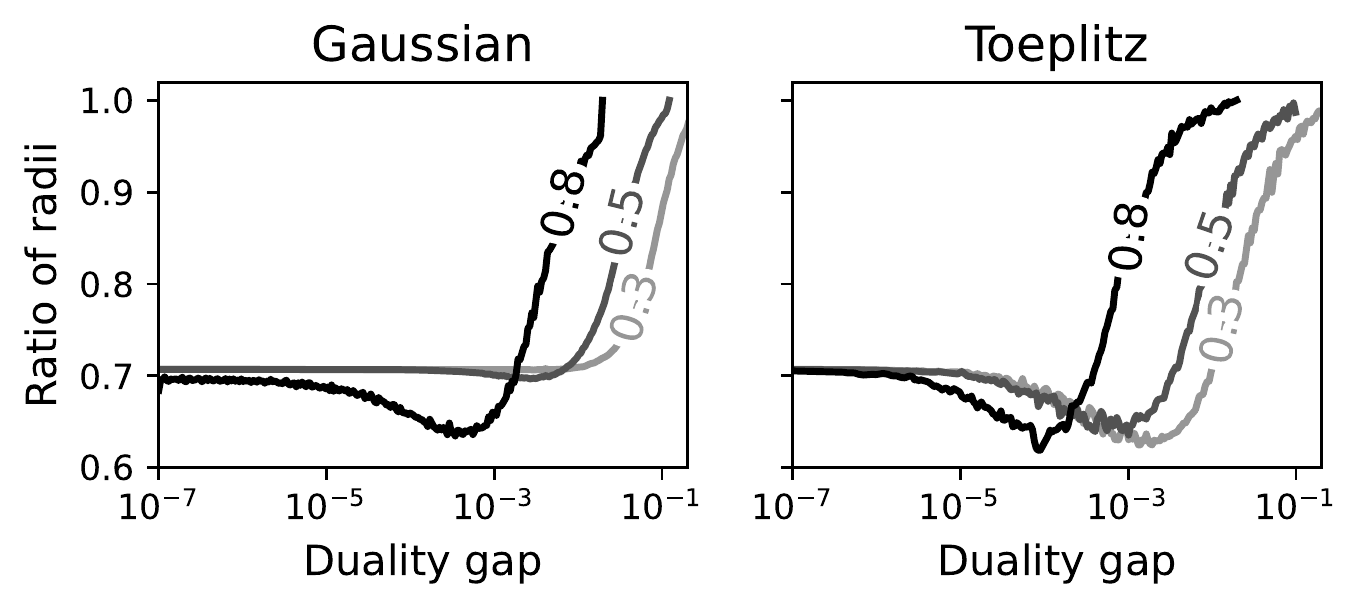}}
	\caption{
		\label{fig: xp radii}
		Expected value of the ratio~\eqref{eq:considered ratio radii in xps} as a function of the duality gap achieved by \((\pv,\dv)\) for the two dictionaries.
		Each curve corresponds to a value of the ratio \(\lambda/\lambda_{\max}\) (the value is indicated along the line).
		% \remCE{The figure is new, potentially prettier (one betters the asymptotic convergence) but the simulations are not over. I will update the figure as soon as they're finished.}
	}
\end{figure}

\vspace*{.5em}

\paragraph{Radius of safe regions}
In this first experiment, we investigate the size difference between the Hölder and GAP domes.
More precisely, we evaluate the ratio
\begin{align}
	\label{eq:considered ratio radii in xps}
	\frac{
		\Radius(\NewDome(\pv,\dv))
	}{
		\Radius(\GapDome(\pv,\dv))
	}
\end{align}
for different choices of \(\pv\in\kR^\pvdim\) and \(\dv\in\dfset\), where \(\Radius(\someset)\) denotes the radius of the (closed bounded) set \(\someset\):
% defined as
\begin{align}
	\label{eq:def radius closed bounded set}
	\Radius(\someset) \triangleq
	\max_{\dv, \dv'\in\someset} \ \tfrac{1}{2}\kvvbar{\dv' - \dv}_2
	.
\end{align}
Figure~\ref{fig: xp radii} shows the average value of the ratio~\eqref{eq:considered ratio radii in xps} as a function of the duality gap achieved by \((\pv,\dv)\) for the two considered dictionaries and three values of the ratio \(\lambda /\lambda_{\max}\).
%\addCE{In our simulations, the parameters \((\pv,\dv)\) used to evaluate the ratio~\eqref{eq:considered ratio radii in xps} are the iterates of an iterative solver that addresses~\eqref{eq: primal}.}
% \remCE{En vrai de vrai: ISTA + dual scaling pour \(\dv\). Mais je ne souhaite pas rentrer dans ce niveau de détail pour ne pas brouiller le lecteur, qu'est-ce que tu en penses?}\remCH{oui ok}
%\remCE{clair?}
% \remCH{ca va. Le lecteur peut juste eventuellement se demander comment \((\pv,\dv)\) sont generés ?}
%\remCE{Oui, je me suis posé la même question; je ne me satisfait d'aucune explication pour l'instant, même sans contraintes de place. J'y réfléchie}
Results have been obtained by averaging over 50 trials.
%.
%\remCH{"trials" plutot ? pour utliser le meme terme que precedemment. } Monte-Carlo simulations.
As expected (cf. \Cref{theo: subsets}), the
%\suppCL{value of the ratio is} 
ratios are always lower than \(1\).
One also observes that the radius of the proposed Hölder dome is up to \(0.6\) smaller than the GAP dome.
As far as our simulation is concerned,
all the curves seem to converge to a ratio close to \(0.7\) as the dual gap tends to zero.
%the value of the ratio seems to converge to a common value closed to \(0.7\) as \((\pv,\dv)\) converges to a primal-dual optimal couple.
% We also emphasized two noteworthy outcomes.

% \Cref{theo: subsets}
% More particulariy
% We recall that the radius of a closed bounded set \(\someset\subset\kR^\obsdim\) is defined by~\addCE{[ref]}
% To state the next theorem, we define \textblue{radius} of a closed bounded set \(\someset\subset\kR^\obsdim\) by

\vspace*{.5em}

% \begin{figure*}[ht!]
% 	\centering
% 	% \includegraphics[width=\textwidth]{img_a_exp.png}
% 	\includegraphics[width=.99\textwidth]{figs/Setup3b_Ista.png}
% 	\caption{
% 		\label{fig: xps}
% 		Performance profiles of screening methods on different setups.
% 		% There are $50$ problem instances for each type of dictionary: Gaussian (left column), Uniform (middle column) and Toeplitz (right column).
% 		Three values of $\lambda$ are considered corresponding to $\lambda/\lambda_{\max}$ equal to $0.3$ (top row), $0.5$ (middle row) and $0.8$ (bottom row).
% 		Here $p(\tau)$ is the percentage of  problem instances achieving dual gap smaller than $\tau$.
% 	}
% \end{figure*}

\paragraph{Benchmarks}
In this second experiment, we assess the computational gain obtained with the Hölder dome and GAP-based regions.
To do so, we compare three variants of FISTA --a standard method to address~\eqref{eq: primal}, see \cite{beck2009fast}-- where the iterations are interleaved with screening tests that leverage
\textit{i)}~the GAP sphere,
\textit{ii)}~the GAP dome,
\textit{iii)}~the Hölder dome.
More precisely, at each iteration \(t\), a screening test is carried out with the corresponding safe region obtained with parameters \((\pv^{(t)},\dv^{(t)})\) where \(\pv^{(t)}\) refers to the current iterate and \(\dv^{(t)}\) is obtained by dual scaling of \(\obs -\dicomat\pv^{(t)}\) (see~\cite[Section 3.3]{Laurent-El-Ghaoui:2012qs}).
% The three method will be respectively referred to as \texttt{GAPs}, \texttt{GAPd} and \texttt{Hölder}.
We use the ``Dolan-Moré'' performance profiles~\cite{Dolan2002} to assess the performance of the three methods.

\begin{figure}[t]
	\centering
	\includegraphics[width=.99\columnwidth]{\paperfigs{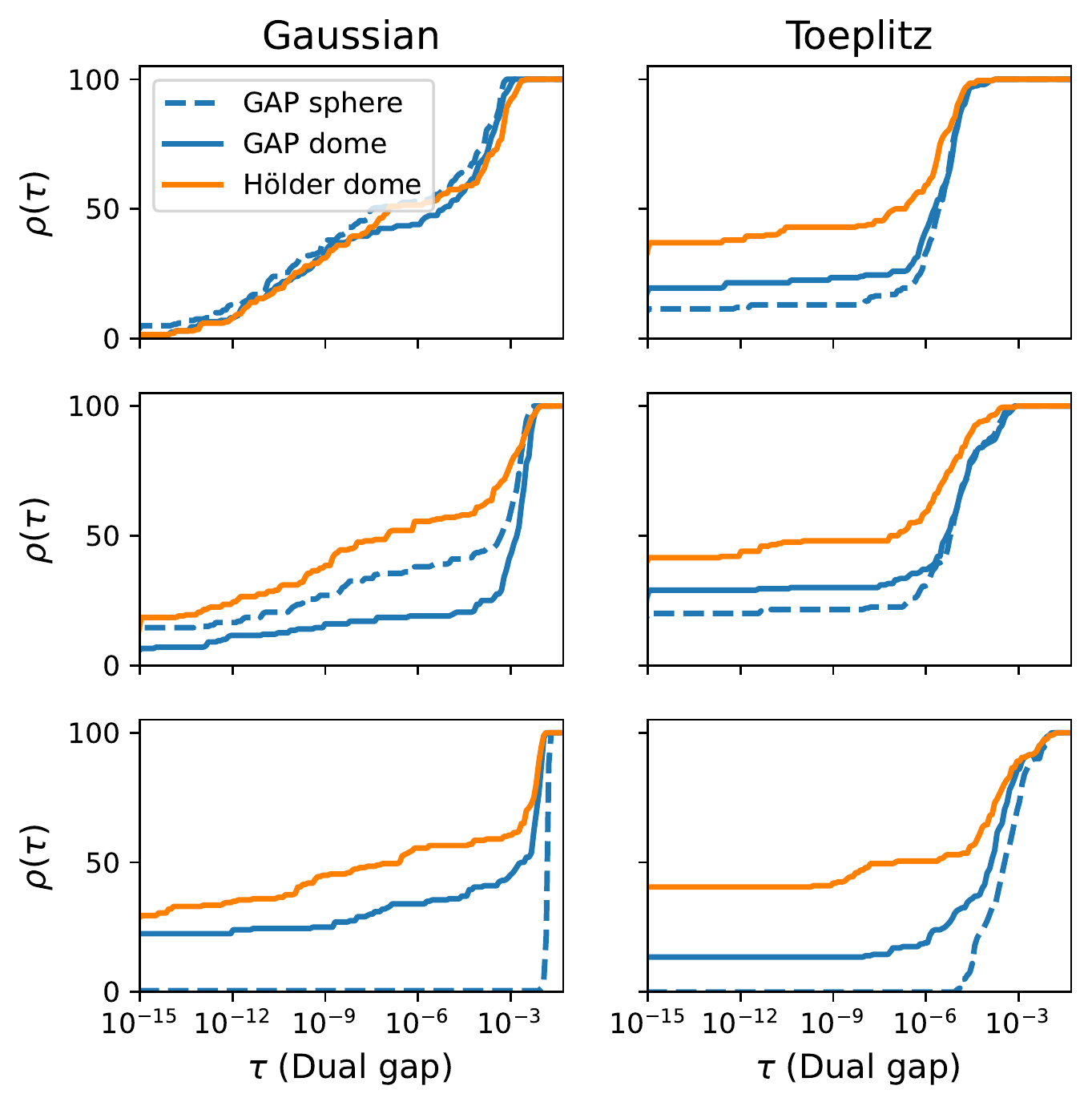}}

	\caption{
		\label{fig: xps}
		Performance profiles of screening methods with different safe regions.
		Each row corresponds to a value of $\lambda/\lambda_{\max}$.
		From top to bottom: $.3,.5,.8$.
	}
\end{figure}

We run each method with a prescribed computational budget (the number of floating point operations) on 200 instances of problem~\eqref{eq: primal}.
We then evaluate the (empirical) probability \(\rho(\tau)\) that a solver achieves a duality gap lower that \(\tau\) upon completion.
For each setup, the budget is adjusted so that \(\rho(10^{-7})=50\%\) for the solver using the Hölder dome.
% for the sake of comparison.
% \remCH{\(\rho_S(\tau)\) sur les figures ?}

Figure~\ref{fig: xps} shows the performance profiles for the two considered dictionaries and different values of the ratio \(\lambda/\lambda_{\max}\).
One sees that, as far as our simulation setup is concerned, implementing the screening test~\eqref{eq:relax screening rule} with the Hölder dome improves quite significantly the average accuracy achieved in all but one setup.
These findings support our claim that the Hölder dome leads to more effective test than the one using GAP based-region.
% Indeed, since the Hölder dome is always a (strict) subset of the GAP dome (see~\Cref{theo: subsets}), one may expect that the screening test are more.
The case where the three safe regions lead to comparable results (Gaussian dictionary and \(\lambda/\lambda_{\max}=0.3\)) has to be understood as follows: even though the Hölder and GAP domes are expected to perform better than the GAP sphere, the profiles result from a compromise between the effectiveness of the test and its complexity.
In particular, a study of our simulation results shows that even though the tests are less effective, more iterations are carried (in average) with the GAP sphere in that specific setup, thus leading to a potentially more accurate solution.

% Hence, one may expect that it allows to detect more zero entries so that each 

% Hence, as far as our simulation is concerned, this shows the

%As far as\dots

%!TEX root = ./paper.tex
%!TEX spellcheck = en_US

\section{Conclusion}\label{section 4}

%\remCE{Je crois que la conclusion n'avait pas été relue}\remCH{En effet, merci :)}

In this paper, we introduced a novel safe dome region for \LASSO{} that can be used to design safe screening tests.
%Our second contribution was to 
We showed that our proposed dome region is always a (potentially strict) subset of the GAP sphere and dome, two ubiquitous safe regions in the literature.
%Finally, we empirically compared the radii of the three safe regions on different setups and demonstrated that 
The proposed methodology is shown to allow significant computational gains when solving \LASSO{} with a prescribed computational budget.\\[-0.2cm]

% In this paper, we introduced a novel dome region for \LASSO{} screening methods.
% Our dome mainly relies on the new cutting half-space $\{\p\in \Rbb^m: \dotp{\A\x}{\p}\leq \lambda \penx\}$ which has normal vector $\A\x$ being the current approximation of observation $\y$.
% Our safe regions converges to the dual optimal solution.
% Moreover, we showed that our NEW dome region is actually a subset of GAP sphere and GAP dome with radius converging to zero faster.
% Numerical experiment show a potential computational gains in comparison with GAP screening.
%Our cutting half-space is applicable for many other types of penalization norm and will be considered in the future works.

% \remCH{Je mets les refs en small: on gagner une demi colonne. Tu epux te faire plez ;)}
% \remCE{Hihi, pas trop quand même, il reste des bouts de preuve :s. Au passage, sans réponse de Le je m'en occupe ce soir.}

\small
\bibliography{references.bib}
\bibliographystyle{IEEEtran}

\normalsize
\appendices

%!TEX root = ./paper.tex
%!TEX spellcheck = en_US

% \vspace{0.2cm}
% \remCE{T'es sur pour le vspace? ca tue le formatage}
\section{Proofs}

\subsection{Proof of Lemma~\ref{lemma: canonical characterization}}
\label{proof:lemma: canonical characterization}
Let $\halfspacen\in\kR^\obsdim$ and consider the following optimization problem
\begin{align}
	\halfspacetreshold^\star =  \sup_{\dv\in\dfset} \ \dotp{\halfspacen}{\dv}.
\end{align}
%\remCL{Should we use $\dotp{\halfspacen}{\dv}$ or $|\dotp{\halfspacen}{\dv}|$? Since (22) uses absolute value} \remCH{It is $\dotp{\halfspacen}{\dv}$. I fixed (22). }
By strong duality
%\remCL{citation request}
\cite[Equation~(12.4) combined with Theorem A.1]{Beck2017aa}, we have %and consider the following problem
\begin{align}\label{eq:dual problem max inner product}
	%\sup_{\dv\in\dfset} \dotp{\halfspacen}{\dv}
	\halfspacetreshold^\star
	 & =
	\inf_{\pv\in\kR^\pvdim} \lambda \|\pv\|_1 + \indcatorfunc\{\halfspacen=\dicomat\pv\}
	,
\end{align}
where $\indcatorfunc\{\cdot\}$ denotes the indicator function, which is equal to \(0\) if the statement in the braces is true and $+\infty$ otherwise.

Hence, if $\halfspacen=\dicomat\pv$ for some $\pv\in\kR^\pvdim$, we have from \eqref{eq:dual problem max inner product} that $\halfspacetreshold^\star\leq \lambda\|\pv\|_1$ and therefore $(\dicomat\pv,\halfspacetreshold)\in\mathcal{G}$ for all $\halfspacetreshold\geq \lambda\|\pv\|_1$.
Conversely, if $(\halfspacen,\halfspacetreshold)\in\mathcal{G}$ then $\halfspacetreshold^\star\leq \halfspacetreshold<\infty$.
From \eqref{eq:dual problem max inner product}, we thus have that there exists some $\pv$ such that $\halfspacen=\dicomat\pv$ and $\halfspacetreshold\geq \lambda\|\pv\|_1$.\\

\subsection{Proof of Theorem~\ref{th:def and safeness new dome}}\label{proof:th:def and safeness new dome}

First remember that a dome is defined as the intersection of the ball $\spheresymb(\spherec,\spherer)$ and the half-space $\halfspacesymb(\halfspacen,\halfspacesymb)$, see \eqref{eq: general def dome as intersection sphere and halfspace}. It is then sufficient to show that both $\spheresymb(\spherec,\spherer)$  and $\halfspacesymb(\halfspacen,\halfspacetreshold)$ are safe. Finally, the safeness of $\spheresymb(\spherec,\spherer)$ and $\halfspacesymb(\halfspacen,\halfspacesymb)$ respectively follows from \cite[Section 2.2]{icml2014c2_liuc14} and Lemma~\ref{lemma: canonical characterization}.\\
% \hfill$\square$\\

\subsection{Proof of Theorem~\ref{theo: subsets}}
\label{sec:app:proof theorem subsets}

%Let \(\pv\in\kR^\pvdim\backslash\{\bf0_\pvdim\}\) and \(\dv\in\dfset\) be a dual feasible point.
%We already show below the statement of \Cref{theo: subsets} that \(\dvopt\in\NewDome(\pv,\dv)\) as a consequence of the Hölder inequality.

%We now turn to the proof of the two inclusions.
% Recall first that the center and radius of the including ball \(\InBall(\dv)\) are defined as
% \(\InBallc=(\obs+\dv)/2\) and \(\InBallr=\kvvbar{\obs - \dv}_2/2\).

% and do not recall the proof here for the sake of concision.
% We now turn to the proof of the two inclusions
% \begin{equation}
% 	\NewDome(\pv,\dv)
% 	\subset \GapDome(\pv,\dv)
% 	\subset \GapBall(\pv,\dv)
% 	.
% \end{equation}

\vspace*{.5em}

%\paragraph{Proof that \(\NewDome(\pv,\dv)\subseteq \GapDome(\pv,\dv)\)}
%Let $\newHalfspace(\pv,\dv)$ (resp. $\GapHalfspace(\pv,\dv)$) denote the half-space defining $\NewDome(\pv,\dv)$ (resp. $\GapDome(\pv,\dv)$). 
Let $\newHalfspace(\pv,\dv)$ and $\GapHalfspace(\pv,\dv)$ respectively denote the half-space defining $\NewDome(\pv,\dv)$ and $\GapDome(\pv,\dv)$.
Consider
\begin{equation}
	\dv'\in\NewDome(\pv,\dv)
	=\spheresymb(\spherec,\spherer)
	\cap \newHalfspace(\pv, \dv),
\end{equation}
where \(\spherec,\spherer\) are the parameters
%of the GAP dome and are
defined in~\eqref{eq: gap dome: spherec} and~\eqref{eq: gap dome: spherer}, respectively.
%\(\dv'\in\NewDome(\pv,\dv)\).
Since $\NewDome(\pv,\dv)$ and $\GapDome(\pv,\dv)$ only differ in the definition of their half-space, it is sufficient to show that $\dv'\in \GapHalfspace(\pv,\dv)$, \ie
% that is 
\begin{align}\label{eq:u' in Hgap}
	\dotp{\obs - \spherec}{\dv'}
	% 	\kangle{
	% 	\GapHalfspacen, \dv' - \InBallc
	% }
	\leq \dotp{\obs - \spherec}{\spherec}+ \gapfun(\pv,\dv)-\spherer^2
	.
\end{align}
% <<<<<<< HEAD
% % \remCE{Comme tous les paramètres des régions safe ont le même notation, je propose d'au moins rappeler de qui on parle ici:}
% where \(\spherec,\spherer\) are the parameters of the GAP dome defined in~\eqref{eq: gap dome: spherec} and~\eqref{eq: gap dome: spherer}.
% =======
% \remCE{Comme tous les paramètres des régions safe ont le même notation, je propose d'au moins rappeler de qui on parle ici:}
% %\addCE{where \(\spherec,\spherer\) are the parameters of the GAP dome and are defined in~\eqref{eq: gap dome: spherec} and~\eqref{eq: gap dome: spherer}, respectively.}
% >>>>>>> 7353c3c063b60c367b6cb56300026bb74ac8722a
% \obs - \spherec
% \dotp{\halfspacen}{\spherec}+ \gapfun(\pv,\dv)-\spherer^2
% By definition of \(\NewDome(\pv,\dv)\), we already have that \(\dv'\in\InBall(\dv)\).
% Let us now show that \(\dv'\in\GapHalfspace(\pv,\dv)\).
We have
\begin{align}
	2\kangle{
		\obs - \spherec, \dv' - \spherec
	}
	\,=\,    &
	%\tfrac{1}{2}\kparen{
	\kvvbar{\obs - \spherec}_2^2
	+ \kvvbar{\dv' - \spherec}_2^2
	- \kvvbar{\obs - \dv'}_2^2
	%}
	\nonumber  \\
	\,\leq\, &
	2\kparen{
		\spherer^2 - \tfrac{1}{2}\kvvbar{\obs - \dv'}_2^2
	}
	\nonumber  \\
	\,=\,    &
	2\kparen{
		\dualfun(\dv') - \dualfun(\dv) - \spherer^2
	}
	\label{eq:proof subsets:part1 first inequality}
\end{align}
% \begin{align}
% 	2\kangle{
% 		\GapHalfspacen, \dv' - \InBallc
% 	}
% 	\,=\,    &
% 	%\tfrac{1}{2}\kparen{
% 	\kvvbar{\obs - \InBallc}_2^2
% 	+ \kvvbar{\dv' - \InBallc}_2^2
% 	- \kvvbar{\obs - \dv'}_2^2
% 	%}
% 	\nonumber  \\
% 	%
% 	\,\leq\, &
% 	2\kparen{
% 		\InBallr^2 - \tfrac{1}{2}\kvvbar{\obs - \dv'}_2^2
% 	}
% 	\nonumber  \\
% 	\,=\,    &
% 	2\kparen{
% 		\InBallr^2 + \dualfun(\dv') - \tfrac{1}{2}\kvvbar{\obs}_2^2
% 	}
% 	\nonumber  \\
% 	\,=\,    &
% 	2\kparen{
% 		\dualfun(\dv') - \dualfun(\dv) - \InBallr^2
% 	}
% 	\label{eq:proof subsets:part1 first inequality}
% \end{align}
where the inequality
%in the second line 
follows from the fact that both \(\obs\) and \(\dv'\) belong to
$\spheresymb(\spherec,\spherer)$.
%\(\InBall(\dv)\).
%We now show that \(\dualfun(\dv') \leq \primalfun(\pv)\).
Moreover,
%using the (trivial) identity \(\obs=\obs-\dicomat\pv+\dicomat\pv\),
one can write
\begin{align}
	2\dualfun(\dv')
	\,=\,    &
	\kvvbar{\obs - \dicomat\pv}_2^2
	- \kvvbar{\obs - \dv' - \dicomat\pv}_2^2
	+ 2\langle\dicomat\pv,\dv'\rangle
	\nonumber  \\
	\,\leq\, &
	\kvvbar{\obs - \dicomat\pv}_2^2
	+ 2\langle\dicomat\pv,\dv'\rangle
	\nonumber  \\
	\,\leq\, &
	\kvvbar{\obs - \dicomat\pv}_2^2
	+ 2\lambda\kvvbar{\pv}_1
	= 2\primalfun(\pv)
	\label{eq:proof subsets:part1 second inequality}
\end{align}
%where the first inequality results from the non-negativity of \(\Vert\obs - \dv' - \dicomat\pv\Vert_2^2\) and the second holds because \(\dv'\in\newHalfspace(\pv,\dv)\) by hypothesis.
where the inequalities result from the non-negativity of \(\Vert\obs - \dv' - \dicomat\pv\Vert_2^2\) and the hypothesis that \(\dv'\in\newHalfspace(\pv,\dv)\).
Combining~\eqref{eq:proof subsets:part1 first inequality} and~\eqref{eq:proof subsets:part1 second inequality}, we finally obtain
%that \eqref{eq:u' in Hgap} holds
\eqref{eq:u' in Hgap},
that is \(\dv'\in\GapHalfspace(\pv,\dv)\).

\iftoggle{techreport}{
	Assume now that \(\primalfun(\pv)<\primalfun({\bf0}_\pvdim)\) and \((\pv, \dv)\) is not primal-dual optimal. We next show that the strict inclusion $\NewDome(\pv,\dv)\subset\GapDome(\pv,\dv)$ holds in that case.
	To this end, it is sufficient to identify some \(\dpoint\in\kR^\obsdim\) satisfying
	\begin{align}
		\dpoint\in \GapDome(\pv, \dv)\setminus\NewDome(\pv, \dv)
		. \label{proof: strict inclusion: goal}
	\end{align}
	% \begin{align}
	% 	\dpoint \triangleq \spherec+\frac{1}{\spherer^2}(\gapfun(\pv,\dv)-\spherer^2)(\obs - \spherec)\in\GapDome(\pv, \dv)\setminus\NewDome(\pv, \dv)	
	% \end{align}
	Let
	%us show that 
	\(\dpoint \triangleq \spherec+\frac{1}{\spherer^2}(\primalfun(\pv)-\dualfun(\dv)-\spherer^2)(\obs - \spherec)\).
	% verifies \eqref{proof: strict inclusion: goal}. 
	We note that \(\dpoint\) is well defined since \(\spherer\neq0\) by construction.
	% \(\dpoint\in \GapDome(\pv, \dv)\setminus\NewDome(\pv, \dv)\).	

	We can show that $\dpoint\in\GapDome(\pv, \dv)$ as follows.
	We first easily verify that
	% =======
	% Assume now that \(\primalfun(\pv)<\primalfun({\bf0}_\pvdim)\) and that \((\pv, \dv)\) is not an optimal primal dual solution.
	% We prove that the inclusion is strict.
	% To this end, we show the existence of a point \(\dpoint\in \GapDome(\pv, \dv)\setminus\NewDome(\pv, \dv)\).
	% %To this end, we construct a point $\dpoint\in \GapDome(\pv, \dv)\setminus\NewDome(\pv, \dv)$.

	% Let \(\dpoint \triangleq \spherec+\frac{1}{\spherer^2}(\primalfun(\pv)-\dualfun(\dv)-\spherer^2)(\obs - \spherec)\).
	% Note that \(\dpoint\) is well defined since \(\spherer\neq0\) by construction.
	% Then, \(\dpoint\in \GapDome(\pv, \dv)\).
	% % \remCE{Requires \(\spherer\neq0\)}
	% Indeed, one easily verifies that
	% >>>>>>> 7353c3c063b60c367b6cb56300026bb74ac8722a
	%it is easy to verify that
	\begin{equation}\label{eq: p0 first property}
		\langle \obs-\spherec, \dpoint-\spherec\rangle
		= \primalfun(\pv)-\dualfun(\dv)-\spherer^2
		.
	\end{equation}
	The result is then proved if $\dpoint\in\spheresymb(\spherec,\spherer)$, \ie
	\begin{equation}
		\label{eq: p0 second property}
		%\kvvbar{\dpoint-\spherec}_2=
		\kvbar{
			\primalfun(\pv)-\dualfun(\dv)-\spherer^2
		}
		< \spherer^2
		.
	\end{equation}
	We can show that~\eqref{eq: p0 second property} holds by
	distinguishing between two cases.
	First, if \(\primalfun(\pv)-\dualfun(\dv) \geq \spherer^2\), the absolute value in~\eqref{eq: p0 second property} can be removed and the result follows from
	%by using the fact that 
	\(\primalfun(\pv)<\primalfun({\bf0}_\pvdim)=\dualfun(\dv)+2\spherer^2\).
	Second, if \(\primalfun(\pv)-\dualfun(\dv) < \spherer^2\), the same conclusion holds by using the fact \(\primalfun(\pv)-\dualfun(\dv)>0\) since \((\pv,\dv)\) is not primal-dual optimal by hypothesis.

	Let us finally show that $\dpoint\notin \NewDome(\pv,\dv)$.
	We have
	\begin{align}
		% \label{eq: p0 result}
		% \begin{aligned}
		2\dualfun(\dpoint)
		 & \,=\, \kvvbar{\obs}_2^2 - \kvvbar{\obs-\dpoint}_2^2
		\nonumber                                              \\
		 & \,=\, \kvvbar{\obs}_2^2
		- \kvvbar{\obs-\spherec}_2^2
		+ 2\langle \obs-\spherec, \dpoint-\spherec\rangle
		- \kvvbar{\dpoint-\spherec}_2^2
		\nonumber                                              \\
		 & \,>\,
		\kvvbar{\obs}_2^2
		- \spherer^2
		+2(\primalfun(\pv)-\dualfun(\dv)-\spherer^2) - \spherer^2
		\nonumber                                              \\
		 & \,=\, 2\primalfun(\pv)
		\nonumber
		%\label{eq: p0 result}
		% \end{aligned}
	\end{align}
	%<<<<<<< HEAD
	where the inequality follows from~\eqref{eq: p0 first property} and~\eqref{eq: p0 second property}. Hence \(\dpoint\notin\NewDome(\pv,\dv)\) since we have from \eqref{eq:proof subsets:part1 second inequality} that any $\dv'\in \NewDome(\pv,\dv)$ must verify $\dualfun(\dv')\leq \primalfun(\pv)$.

	%otherwise \eqref{eq:proof subsets:part1 second inequality} implies that $\dualfun(\dpoint)\leq \primalfun(\pv)$ which is in contradiction with~\eqref{eq: p0 result}.

	%that is the following inequality holds:
}{ Finally,
	%if
	% \(\primalfun(\pv)<\primalfun({\bf0}_\pvdim)\) and \((\pv, \dv)\) is not primal-dual optimal, the strict inclusion $\NewDome(\pv,\dv)\subset\GapDome(\pv,\dv)$ can be seen by noticing that 
	%\(\dpoint \triangleq \spherec+\frac{1}{\spherer^2}(\primalfun(\pv)-\dualfun(\dv)-\spherer^2)(\obs - \spherec)\in\GapDome(\pv,\dv)\) but does not belong to $\NewDome(\pv,\dv)$, see \cite[Appendix~\ref{techreport-proof:eq:inclusion sage GAP regions}]{Le2022_HolderDomeTechreport}.
	if \(\primalfun(\pv)<\primalfun({\bf0}_\pvdim)\) and \((\pv, \dv)\) is not primal-dual optimal,
	then \addCE{one can verify that} \(\dpoint \triangleq \spherec+\frac{1}{\spherer^2}(\primalfun(\pv)-\dualfun(\dv)-\spherer^2)(\obs - \spherec)\) belongs to \(\GapDome(\pv,\dv)\) but does not belong to $\NewDome(\pv,\dv)$ \addCE{(see \cite[Appendix~\ref{techreport-proof:eq:inclusion sage GAP regions}]{Le2022_HolderDomeTechreport} for more details)}, \ie the strict inclusion $\NewDome(\pv,\dv)\subset\GapDome(\pv,\dv)$ holds true.

	% (see \iftoggle{techreport}{Appendix~\ref{proof:eq:inclusion sage GAP regions}}{\cite[Appendix~\ref{techreport-proof:eq:inclusion sage GAP regions}]{Le2022_HolderDomeTechreport}})
}

\vspace*{.5em}

\iftoggle{techreport}{

	\section{Proof of~\eqref{eq:inclusion sage GAP regions}}\label{proof:eq:inclusion sage GAP regions}

	%\paragraph{\(\GapDome(\pv,\dv)\subseteq \GapBall(\pv,\dv)\)}
	Let \(\dv'\in\GapDome(\pv,\dv)\).
	By definition, we thus have
	\begin{align}
		\norm{\dv'- \spherec}_2           & \leq \spherer\label{eq:sphere constraint}                        \\
		\dotp{\halfspacen}{\dv'-\spherec} & \leq \gapfun(\pv,\dv)-\spherer^2\label{eq:half-space constraint}
	\end{align}
	where \(\spherec\), \(\spherer\) and \(\halfspacen\) are the parameters of the GAP dome defined in~\eqref{eq: gap dome: spherer}-\eqref{eq: gap dome: halfspacen}.
	We want to show that
	\begin{align}
		\|\dv'-\dv\|_2 \leq \sqrt{2 \gapfun(\pv,\dv)}.
	\end{align}
	%Remembering that \(\GapBallc=\dv\), we have
	This follows from
	\begin{align}
		% \norm{\dv' - \GapBallc}_2^2
		%
		\norm{\dv' - \dv}_2^2
		% \nonumber                     \\
		 & = \norm{\dv'-\spherec}_2^2
		+2\dotp{\spherec-\dv}{\dv'-\spherec}
		+\norm{\spherec-\dv}_2^2
		\nonumber                     \\
		 & \leq  \spherer^2
		+ 2\, \kparen{\gapfun(\pv,\dv)
			% + 2\kparen{
			% 	\primalfun(\pv) - \dualfun(\dv)
			- \spherer^2
		} + \spherer^2
		\nonumber                     \\
		 & = 2\, \gapfun(\pv,\dv)
		%  \kparen{
		% 	\primalfun(\pv) - \dualfun(\dv)
		% }
	\end{align}
	where the inequality follows from \eqref{eq:sphere constraint}, \eqref{eq:half-space constraint} and the fact that \(\spherec-\dv=\obs-\spherec\).

}{
	% Empty otherwise
}

\end{document}